\documentclass[letterpaper, 10 pt, journal, twoside]{IEEEtran} 

\IEEEoverridecommandlockouts                              

\usepackage{graphicx} 
\usepackage{amsmath} 
\usepackage{amssymb}  
\usepackage{color,soul}
\usepackage{comment}
\usepackage[normalem]{ulem}
\newcommand{\revision}{\textcolor{black}}
\newcommand{\secondrevision}{\textcolor{black}}
\newcommand{\tom}{\textcolor{black}}
\setstcolor{red}
\newcommand{\previous}{\textcolor{red}}
\newcommand{\fakest}[1]{\unskip}
\newcommand{\secondfakest}[1]{\unskip}
\title{KNODE-MPC: A Knowledge-based Data-driven Predictive Control Framework for Aerial Robots
}

\author{Kong Yao Chee$^*$, Tom Z. Jiahao$^*$ and M. Ani Hsieh
\thanks{Manuscript received: September 9, 2021; Revised: December 3, 2021; Accepted: December 27, 2021.}
\thanks{This paper was recommended for publication by Editor C. Gosselin upon evaluation of the reviewers' comments. 
This work was supported by NSF IIS 1910308 and DSO National Laboratories, 12 Science Park Drive, Singapore 118225. $^*$\textit{(Kong Yao Chee and Tom Z. Jiahao are co-first authors.) (Corresponding author: Kong Yao Chee)}}
\thanks{The authors are with the GRASP Laboratory, University of Pennsylvania, Philadelphia, PA 19104, USA.
        {\tt\footnotesize \{ckongyao,\, zjh,\,m.hsieh\}@seas.upenn.edu}}
\thanks{Digital Object Identifier (DOI): see top of this page.}
}

\begin{document}

\markboth{IEEE Robotics and Automation Letters. Preprint Version. Accepted December, 2021}
{Chee \MakeLowercase{\textit{et al.}}: KNODE-MPC}

\maketitle

\begin{abstract}
In this work, we consider the problem of deriving and incorporating accurate dynamic models for model predictive control (MPC) with an application to quadrotor control. MPC relies on precise dynamic models to achieve the desired closed-loop performance. However, the presence of uncertainties in complex systems and the environments they operate in poses a challenge in obtaining sufficiently accurate representations of the system dynamics. In this work, we make use of a deep learning tool, knowledge-based neural ordinary differential equations (KNODE), to augment a model obtained from first principles. The resulting hybrid model encompasses both a nominal first-principle model and a neural network learnt from simulated or real-world experimental data. Using a quadrotor, we benchmark our hybrid model against a state-of-the-art Gaussian Process (GP) model and show that the hybrid model provides more accurate predictions of the quadrotor dynamics and is able to generalize beyond the training data. To improve closed-loop performance, the hybrid model is integrated into a novel MPC framework, known as KNODE-MPC. Results show that the integrated framework achieves \previous{\fakest{73\%}}\revision{60.2\%} improvement in simulations and more than \previous{\fakest{14\%}}\revision{21\%} in physical experiments, in terms of trajectory tracking performance.
\end{abstract}


\section{Introduction}
\IEEEPARstart{T}{he} advent of model predictive control (MPC) has enabled control processes to take advantage of the rapid advancement in optimization techniques and the computational power of modern hardware \cite{qin2003survey}. The ability to optimize over  \revision{accurate} dynamic models and incorporate various constraints has bestowed model predictive controllers with feasibility and stability guarantees for complex control tasks \cite{borrelli2017predictive}. However, identifying accurate dynamic models remains a central challenge to MPC. Building accurate first-principle models often requires domain expertise and deep physical insights into the systems. Even for experts, modeling can be a daunting and tedious task when the system or its operating environment is complex. 

Recent advances in machine learning algorithms have enabled efficient discovery of coherent patterns in complex data. Neural networks (NNs) have been used for various tasks in computer vision, natural language processing, and recommender systems. Most recently, a new family of neural networks -- neural ordinary differential equations (NODE), has been shown as an effective tool for extracting dynamic models from data. NODE approximates a continuous-depth neural network which is used to directly model differential equations. Knowledge-based NODE (KNODE) was proposed to leverage NODE's compatibility with first principles. It combines first-principle models and NNs into hybrid models to improve the open-loop prediction accuracy of nonlinear dynamic models \cite{jiahao2021knowledgebased}. However, it is unclear whether such hybrid models could improve the closed-loop performance of robotic systems in the real world.

\begin{figure}
    \centering
    {\includegraphics[scale=0.295]{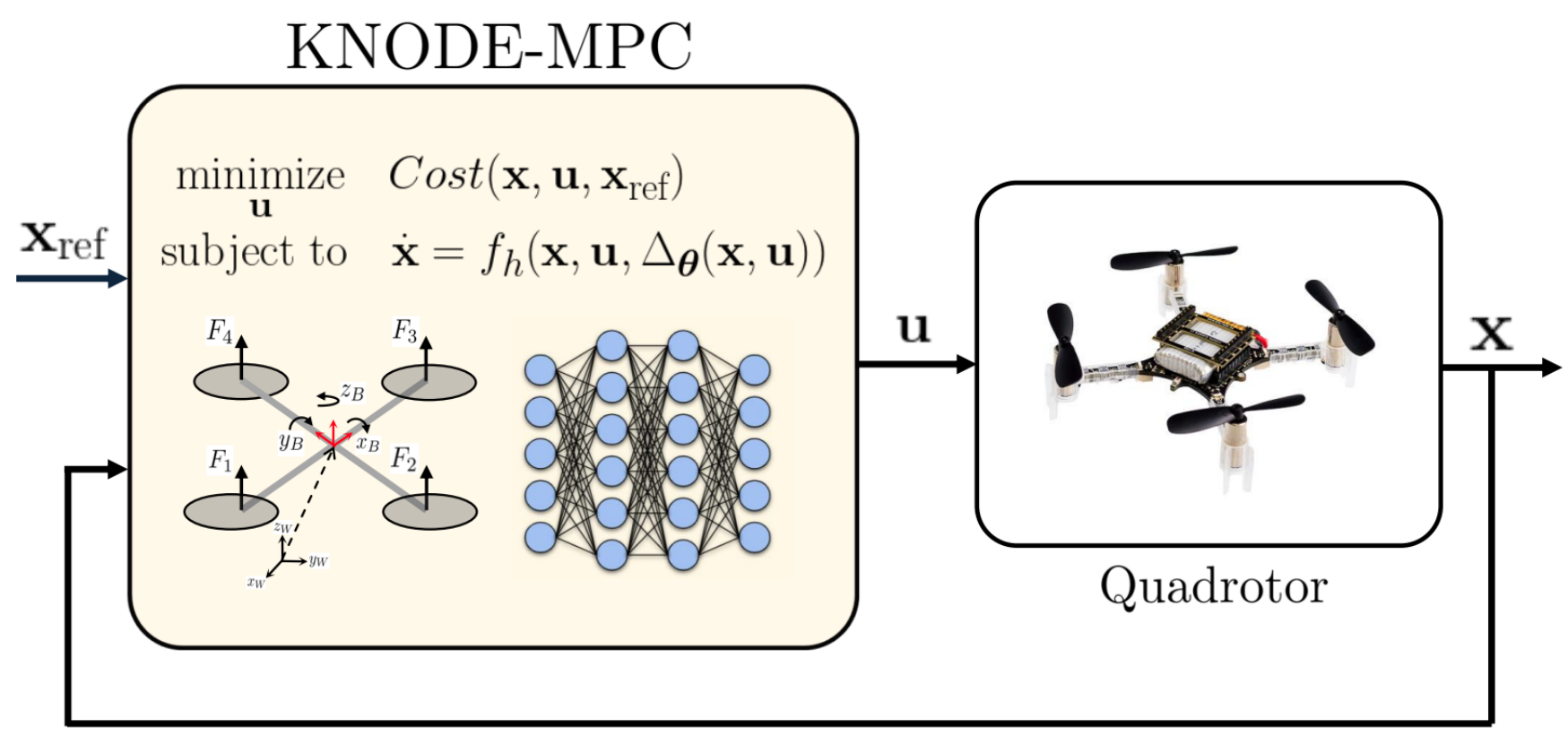}}
    \caption{Schematic of the KNODE-MPC framework, applied to a quadrotor system. The NN accounts for residual or uncertain dynamics in the system, and is combined with a model derived from first principles to form a KNODE model. The KNODE model is then incorporated into a MPC framework, where optimal control commands are generated and applied to the quadrotor.}
    \label{fig:cover_image}
\end{figure}

In this work, we use KNODE to model the dynamics of a quadrotor. In particular, we utilize the neural network in the hybrid model to account for residual and uncertain dynamics within the system. This hybrid model is then incorporated into a model predictive control framework, known as KNODE-MPC. A schematic of the proposed framework is depicted in Fig. \ref{fig:cover_image}. We demonstrate that the integrated framework not only provides better predictions of future states, but also improves the closed-loop performance in both simulations and real-world experiments. 

\section{Related Work}
In recent years, increasing data availability has fueled rapid development in data-driven system identification. One body of work aims to extract the governing equations of dynamic models directly from data \cite{SINDy, EntropicRegression, Champion22445}. A major contribution uses sparse regression to reconstruct dynamic models from a predetermined library of basis functions \cite{SINDy}. This method has been shown as an effective alternative to system identification \cite{Kaiser_2018}. However, a fundamental challenge to this method lies in coming up with the library of functions. If the correct basis functions are missing from the library, the sparse regression fails to identify the correct dynamic model. Such methods also do not scale well to high-dimensional systems as the library of functions can become unreasonably large. This prevents deployment on real-world systems, which are very often high-dimensional. 

On the other hand, neural networks have recently been shown as an effective parameterization of dynamic models. Stochastic neural networks were used to model residual physics in contact models \cite{ajay2018augmenting}, and subsequently ContactNets was proposed to model frictional contact behaviors \cite{pfrommer2020contactnets}. Recurrent neural networks (RNNs) have been used to model the dynamics of high-dimensional flows \cite{qraitem2020bridging} and spatio-temporally chaotic systems \cite{OttRescomp}. NODE was developed for similar purposes but differs from RNNs by explicitly modeling differential equations from data \cite{conf_nips_ChenRBD18}. Notably, NODE has enabled various deep learning tools to facilitate system identification. For example, convolutional neural networks have been used to reduce the dimensionality of high-dimensional systems \cite{jiahao2021knowledgebased}. In particular, KNODE has demonstrated neural network's generality by modeling a variety of systems with nonlinear and chaotic dynamics, and has also shown NODE's compatibility with first-principle dynamic models \cite{jiahao2021knowledgebased}. These developments have opened up new avenues for system identification for MPC.

There are a number of works in the literature that integrate data-driven models into a predictive control framework. One approach to synthesizing models for MPC is through Gaussian processes (GP) regression. In \cite{kabzan2019learning}, GP regression is used to learn the residual errors between the true and nominal dynamics of a race car, accounting for prior knowledge of the system. The authors in \cite{torrente2021data} extend this framework to model residual dynamics in quadrotors. However, these methods assume that the relationship between the residual and true dynamics is known and this limits the nature of uncertainties that these models are able to accommodate. Another drawback of GP regression is computational complexity. This implies the need to select a small number of training data points that best represent the dynamics. Since the dynamics are either unknown or uncertain, it is difficult to select these points in practice. Another increasingly popular approach that incorporates learning into MPC is model-based reinforcement learning (MBRL). In \cite{williams2017information}, neural networks are used to learn a dynamics model, which is then applied to a MPC framework to solve iterative tasks. The authors in \cite{lambert2019} use MBRL to design low-level controllers using sensors on-board the quadrotor. Compared to the approaches that use GP regression and our proposed framework, these methods do not account for prior knowledge of the system dynamics and are therefore less sample-efficient. 

Our contributions in this work are three-fold. First, we employ KNODE to develop a high-fidelity quadrotor model capable of capturing poorly understood uncertainties and residual dynamics. The KNODE model combines prior knowledge of the system dynamics with a neural ODE, and provides a better representation of the quadrotor dynamics. Second, we show that the KNODE model improves the accuracy of state predictions significantly, over both the nominal model, as well as a state-of-the-art GP-based prediction model, which is similar to those described in \cite{kabzan2019learning} and \cite{torrente2021data}. In particular, we compare our framework against a more general variant of these GP models that does not impose any structural assumptions between the system uncertainties and the true dynamics. Third, the hybrid model is integrated into a novel model predictive control framework, known as KNODE-MPC. Simulations and physical experiments are conducted with KNODE-MPC to evaluate the improvement in closed-loop performance. Empirical results show that the framework outperforms a nominal model predictive controller with a first-principle model.

\section{Methodology}

\subsection{Quadrotor Dynamics} 
In this work, we use a six degrees-of-freedom quadrotor system to illustrate our proposed KNODE-MPC framework. The model used to describe this system is similar to that in \cite{mellinger2011}. Denoting the mass and moments of inertia as $m$ and $J := \text{diag}(J_{xx},\,J_{yy},\,J_{zz})$, the dynamics of the quadrotor, governed by its equations of motion, can be described as:
\begin{equation} \label{eq:forces}
    \begin{bmatrix} \ddot{x}\\\ddot{y}\\\ddot{z} \end{bmatrix} := \ddot{\mathbf{r}} = \frac{1}{m} \left( \begin{bmatrix} 0\\0\\-mg \end{bmatrix} + R_B^W \begin{bmatrix} 0\\0\\u_1 \end{bmatrix} \right),\\
\end{equation}
\begin{equation} \label{eq:moments}
    \begin{bmatrix} \dot{p}\\\dot{q}\\\dot{r} \end{bmatrix} := \dot{\boldsymbol{\omega}} = J^{-1} \left( \mathbf{u}_2 - \begin{bmatrix} p\\q\\r \end{bmatrix} \times J \begin{bmatrix} p\\q\\r \end{bmatrix} \right),
\end{equation}
where $R_B^W$ is the transformation matrix between the quadrotor body frame and the world frame. $u_1$ is the summation of the motor forces, \textit{i.e.}, $u_1 := \sum_{i=1}^4 F_i$ and $\mathbf{u}_2$ is the moment vector that is related to the motor forces by
\begin{equation} \label{eq:motor_moments}
    \mathbf{u_2} = \begin{bmatrix} 0 & L & 0 & -L\\ -L & 0 & L & 0\\ \gamma & -\gamma & \gamma & -\gamma \end{bmatrix} \begin{bmatrix} F_1 \\ F_2 \\ F_3 \\ F_4 \end{bmatrix},
\end{equation}
where $L$ is the arm length of the quadrotor and $\gamma$ is the ratio between the moment and thrust coefficient of the motors. The orientation in terms of Euler angles velocities is related to the angular velocities by
\begin{equation} \label{eq:orientation}
    \boldsymbol{\omega} = \begin{bmatrix} cos(\theta) &0 &-\cos(\phi)\sin(\theta) \\0 &1 &\sin(\phi) \\
    \sin(\theta) & 0 & \cos(\phi)\cos(\theta) \end{bmatrix} \begin{bmatrix} \dot{\phi} \\ \dot{\theta} \\ \dot{\psi} \end{bmatrix}.
\end{equation}

By defining the state as $\mathbf{x} := [\mathbf{r}\;\dot{\mathbf{r}}\;\phi\;\theta\;\psi\;\boldsymbol{\omega}] \in \mathbb{R}^n$, we have the following compact representation of the quadrotor system,
\begin{equation} \label{eq:nominal}
    \dot{\mathbf{x}} = f(\mathbf{x},\mathbf{u}),
\end{equation}
with $\mathbf{u} = [u_1\; \mathbf{u}_2]^T \in \mathbb{R}^m$ and $f: \mathbb{R}^n \times \mathbb{R}^m \rightarrow \mathbb{R}^n$ denotes a mapping given by \eqref{eq:forces}, \eqref{eq:moments} and \eqref{eq:orientation}. In this work, both quaternions and Euler angles are used interchangeably to represent orientation. 

\subsection{Knowledge-based Neural ODEs} \label{KNODE_section}
Although the nominal quadrotor model in \eqref{eq:nominal} represents the system with certain fidelity, it is generally not sufficient for applications where a more accurate model is required. For instance, the quadrotor experiences aerodynamic and environmental disturbances in flight, and these are not captured by the nominal model. In addition, system parameters such as the moments of inertia are inherently difficult to measure or estimate. These uncertainties cause deviations between the true dynamics of the system and those given by the nominal model. The KNODE framework tackles this challenge by learning these uncertainties and residual dynamics using a data-driven approach. By combining prior knowledge of the system with the learned dynamics, this hybrid model not only improves accuracy, but is also more sample efficient \revision{\cite{jiahao2021knowledgebased}}. 

Consider the following representation of the true system dynamics
\begin{equation} \label{eqn:uncertainDynamics}
    \dot{\mathbf{x}} = f_t(\mathbf{x},\mathbf{u},\, \Delta(\mathbf{x},\mathbf{u})),
\end{equation}
where $f_t : \mathbb{R}^n \times \mathbb{R}^m \times \mathbb{R}^p \rightarrow \mathbb{R}^n$ denotes the true dynamics. The subscript $t$ denotes the true system. $\Delta : \mathbb{R}^n \times \mathbb{R}^m \rightarrow \mathbb{R}^p$ is a mapping from the state-input space to the uncertainty space. Next, we parameterize the uncertainty $\Delta$ using an NN with parameters $\boldsymbol{\theta}$.  This represents part of the true dynamics that are absent in the nominal model. With this parameterization, the hybrid KNODE model (subscript $h$) can be written as
\begin{equation} \label{eqn:hybrid_model_coupled}
    \dot{\mathbf{x}} = f_h(\mathbf{x},\mathbf{u},\, \Delta_{\boldsymbol{\theta}}(\mathbf{x},\mathbf{u})),
\end{equation}
where $\Delta_{\boldsymbol{\theta}}$ is a neural network, and $f_h$ includes both the nominal model and its coupling with the neural network. This framework is general enough to allow coupling between different components of the state and input, and it \previous{\fakest{accommodates}}\revision{is able to accommodate} uncertain forces and moments \previous{\fakest{readily}}. Furthermore, compared to non-parametric methods like Gaussian process (GP) regression, this parametric approach allows training on large amounts of data while keeping the size of the uncertainty model constant, and can potentially be computationally faster during inference. Hence, by modeling uncertainty in the dynamics, this hybrid model gives a more accurate representation of the true system. 

\subsection{Model Predictive Control}

An MPC framework uses an optimization-based approach to generate a sequence of optimal control inputs in a receding horizon manner. In its formulation, a model of the system, together with state and input constraints, are considered within an optimization problem of the following form,
\begin{equation} \label{eq:mpc}
\begin{split}
    \underset{\mathbf{u}}{\textnormal{minimize}}\quad\; & \sum_{i=1}^{N-1} \left( x_i^T Q x_i + u_i^T R u_i \right) + x_N^T P x_N\\
    \text{subject to}\quad\; &x_{i+1} = f_{d}(x_i, u_i), \quad \forall\, i=0,\dotsc,N-1,\\
    \; &x_i \in \mathcal{X}, \quad u_i \in \mathcal{U}, \quad \forall\, i=0,\dotsc,N-1,\\
    \; &x_0 = x(k),
\end{split}
\end{equation}
where $\mathcal{X}$ and $\mathcal{U}$ are sets defining the state and input constraints and $x(k)$ is the initial state to the optimization problem at time step $k$. $Q$, $R$ and $P$ are weighting matrices for the stage, input and terminal costs respectively. At each time step, \eqref{eq:mpc} is solved to obtain a sequence of control inputs $\mathbf{u}:=[u_0,\dotsc,u_{N-1}]^T$ and $u_0$ is applied to the system. The discrete-time model $f_d : \mathbb{R}^n \times \mathbb{R}^m \rightarrow \mathbb{R}^n$ is derived from a continuous-time counterpart using a numerical solver. An explicit fourth-order Runge-Kutta method is applied to obtain the predicted states $x_{i+1},\,i=0,\dotsc,N-1$. In the ideal case, the true dynamics in \eqref{eqn:uncertainDynamics} are discretized, but since the uncertainty is unknown, a discrete-time version of the nominal model described in \eqref{eq:nominal} is commonly used in this framework. 

There exist robust and stochastic MPC methods in the literature that account for uncertainty by assuming that the uncertainty is either bounded \cite{bemporad1999robust,mayne2011tube} or characterized by a probability distribution \cite{mesbah2016a,mesbah2016b}. However, these approaches require additional assumptions on the nature of the uncertainty, which can be challenging to ascertain in practice. In most cases, the closed-loop performance is sensitive to the accuracy of the model and hence, using the KNODE model in \eqref{eqn:hybrid_model_coupled} is expected to yield better closed-loop system performance than a nominal model. This motivates the KNODE-MPC framework, where a hybrid knowledge-based NODE model is integrated into the MPC framework. 

\subsection{KNODE Training}

While the KNODE framework is applicable to the general case shown in \eqref{eqn:uncertainDynamics}, we shall focus on a more practical scenario. Assuming that the uncertainty is separable from the nominal dynamics, the true dynamics can then be expressed as
\begin{equation} 
    \dot{\mathbf{x}} = f_t(\mathbf{x},\mathbf{u},\,\Delta(\mathbf{x},\mathbf{u})) = f(\mathbf{x},\mathbf{u}) + \Delta(\mathbf{x},\mathbf{u}),
\end{equation}
with the first component given by the nominal dynamics in \eqref{eq:nominal}. Following the parameterization described in \ref{KNODE_section}, the hybrid model that consists of both the nominal model and neural network is given by
\begin{equation} \label{eqn:hybrid_model}
    \dot{\mathbf{x}} = f_h(\mathbf{x},\mathbf{u},\Delta_{\boldsymbol{\theta}}(\mathbf{x},\mathbf{u})) = f(\mathbf{x},\mathbf{u}) + \Delta_{\boldsymbol{\theta}}(\mathbf{x},\mathbf{u}),
\end{equation}
where $\Delta_{\boldsymbol \theta}$ denotes the neural network. Training of $\Delta_{\boldsymbol \theta}$ is done with the trajectory data collected from closed-loop simulations or experiments. The training process in our work differs from \cite{jiahao2021knowledgebased} in that control inputs in the training data need to be injected into the model to make the one-step-ahead predictions. Similar to \cite{jiahao2021knowledgebased}, we first divide up trajectory data from each time step, and then use the state at each of these time steps as initial conditions to make one-step-ahead predictions using \eqref{eqn:hybrid_model}. Specifically, we consider a set of state observations and control inputs $\mathcal{O} = [(\mathbf{x}(t_1), \mathbf{u}(t_1)), (\mathbf{x}(t_2), \mathbf{u}(t_2)), \cdots, (\mathbf{x}(t_N), \mathbf{u}(t_N))]^T$, sampled at the times $T = \{t_1, t_2,\cdots,t_N\}$. For a time step $t_i$, we compute the one-step-ahead prediction given by
\previous{\fakest{$\hat{\mathbf{x}}(t_{i+1}) = \mathbf{x}(t_i) +  \int^{t_{i+1}}_{t_i} f_h(\mathbf{x}(t_i),\mathbf{u}(t_i))) dt$,}}
\begin{equation}
    \revision{\hat{\mathbf{x}}(t_{i+1}) = \mathbf{x}(t_i) +  \int^{t_{i+1}}_{t_i}f(\mathbf{x},\mathbf{u}) + \Delta_{\boldsymbol{\theta}}(\mathbf{x},\mathbf{u}) dt},
    \label{eqn: cont one-step-ahead}
\end{equation}
where $\hat{\mathbf{x}}(t_{i+1})$ denotes the state prediction at time $t_{i+1}$. The learning framework in this work performs the integration in \eqref{eqn: cont one-step-ahead} numerically using the explicit fourth order Runge-Kutta method. Next, to compute the mean squared error (MSE) between the predictions and the ground truth, we define a loss function as
\previous{\fakest{
$L(\boldsymbol{\theta}) := \frac{1}{N-1}\sum^{N}_{i=2}(\hat{\mathbf{x}}(t_i) - \mathbf{x}(t_i)),$
}}
\revision{
\begin{equation}
    L(\boldsymbol{\theta}) := \frac{1}{N-1}\sum^{N}_{i=2}\left\|\hat{\mathbf{x}}(t_i) - \mathbf{x}(t_i)\right\|^2_2,
    \label{eqn: ave mse}
\end{equation}
}
and the optimization problem is therefore given by
\previous{\fakest{
$\textrm{s.t.} \quad \dot{\mathbf{x}} = f_h(\mathbf{x},\mathbf{u}, \Delta_{\boldsymbol{\theta}}(\mathbf{x},\mathbf{u})),$
}}
\begin{equation}
\begin{aligned}
\min_{\boldsymbol{\theta}} \quad & L(\boldsymbol{\theta}) \\
\revision{\textrm{s.t.} \quad} & \revision{\dot{\mathbf{x}} = f(\mathbf{x},\mathbf{u}) + \Delta_{\boldsymbol{\theta}}(\mathbf{x},\mathbf{u})},
\end{aligned}
\label{eqn: optimization}
\end{equation}

where the constraint comprises of the dynamic model. To train the hybrid model on multiple trajectories, we can sum up the losses computed on each trajectory. Similar to \cite{jiahao2021knowledgebased, conf_nips_ChenRBD18}, we solve the optimization task in \eqref{eqn: optimization} using the adjoint sensitivity method  which propagates gradients from the loss function to the neural network parameters. The parameters are then iteratively updated based on the gradients using readily available optimizers like Adam \cite{Kingma2015AdamAM}. The adjoint sensitivity method has been noted as a memory-efficient alternative to the conventional backpropagation \cite{conf_nips_ChenRBD18} and its proof can be found in \cite{Cao2003AdjointSA}.

\subsection{Evaluation and Verification} \label{section:evaluation}
In our experimental design and evaluation process, we set out to answer the following questions about our proposed KNODE-MPC framework. (a) How accurate are the state predictions given by the KNODE model, as compared to state-of-the-art data-driven models in \cite{kabzan2019learning} and \cite{torrente2021data}? (b) How well does the KNODE model generalize beyond training data? (c) How much improvement does the KNODE-MPC framework provide, in terms of closed-loop trajectory tracking, compared to \previous{\fakest{a nominal MPC framework}}\revision{both the nominal MPC framework, as well as a MPC framework that uses a GP model?} 

To evaluate the accuracy of state predictions and to verify the generalization capability of the KNODE model, a generalized variant of the model in \cite{torrente2021data} is used as a benchmark. In \cite{torrente2021data}, it is assumed that the relationship between the uncertainties and equations of motion are known and this is incorporated as user-defined selection matrices within the model (see Section \previous{\fakest{III.C}}\revision{III-C} of \cite{torrente2021data}). The model also assumes that the uncertainties are decoupled and only allows for \previous{\fakest{a mapping}}\revision{scalar mappings} between the body velocities and acceleration disturbances \revision{in each of the axes} (see Section \previous{\fakest{III.E}}\revision{III-E} of \cite{torrente2021data})\previous{\fakest{, without accounting for any rotational disturbances}}. These assumptions are often difficult to ascertain in practice. 

In this work, we implement a general version of this GP model that removes these assumptions, which can be compactly represented as
\begin{equation} \label{eq:gp_model} 
    \dot{\mathbf{x}}_{\textbf{gp}} = f_{\textbf{gp}}(\mathbf{x},\mathbf{u}) = f(\mathbf{x},\mathbf{u}) + \boldsymbol \mu \left(\begin{bmatrix}\mathbf{x} \\\mathbf{u}\end{bmatrix}\right),
\end{equation}
where $f(\mathbf{x},\mathbf{u})$ represents the nominal model in \eqref{eq:nominal} and $\boldsymbol{\mu}$ denotes the mean posterior of a GP. A nominal model that does not account for uncertainty is also included in our tests as a baseline. To demonstrate the effectiveness of both data-driven models, we incorporate nonlinear components into the true dynamics and compare the prediction accuracy across three models; the nominal, KNODE and GP models from \eqref{eq:nominal}, \eqref{eqn:hybrid_model} and \eqref{eq:gp_model} respectively. \revision{More details on these nonlinear components are given in Section \ref{section:setup}}. Quantitatively, we want to measure the spatial similarity between the true and predicted trajectories and this can be evaluated using a distance metric generated by the dynamic time warping (DTW) algorithm \cite{sakoe1978}. \revision{The DTW algorithm is implemented using the tslearn package \cite{JMLR:v21:20-091}}. For evaluation of the closed-loop trajectory tracking performance of KNODE-MPC, we compare it against \previous{\fakest{a nominal MPC framework, both in simulations and in physical experiments.}}\revision{two frameworks, a nominal MPC framework, as well as a MPC framework that uses the GP model, which we denote as GP-MPC}. The nominal MPC framework uses a model that does not consider uncertainty such as aerodynamic forces or environmental disturbances in its formulation. The closed-loop trajectories obtained from \previous{\fakest{both frameworks}}\revision{all three frameworks} are compared against the planned trajectory using the same distance metric \revision{and these comparisons are done in both simulations and physical experiments}.

\section{Experiments and Results}

\subsection{Setup} \label{section:setup}
\textit{Simulations}: A nominal quadrotor model is constructed using the equations of motion given in \eqref{eq:nominal}. An explicit 5$^{\text{th}}$ order Runge-Kutta method (RK45) with a sampling interval of 2 milliseconds is used for numerical integration to generate dynamic responses of the quadrotor. We assume that the model predictive controller has access to perfect measurements of the quadrotor dynamics. The MPC architecture is implemented in CasADi \cite{Andersson2019} and the optimization routine generates optimal thrust and moment commands. These commands act as inputs to the quadrotor model, which simulate\previous{\fakest{s}} the closed-loop responses. Two classes of trajectories are considered in the simulations; circular and lemniscate trajectories. Both the GP and KNODE models are trained on circular trajectories of radii 3m and 6m \previous{\fakest{, and a}}\revision{only, without lemniscate trajectories. A} circular trajectory of radius 4m is used as validation data. \revision{These data are collected at a commanded speed of 1m/s.} To model uncertainty within the system, nonlinear aerodynamic drag effects from the fuselage and rotors are incorporated into the true quadrotor dynamics. \revision{Inspired from \cite{torrente2021data}, the drag effects are formulated as
\begin{equation}
    D_{\text{total}}^B = D_{\text{rotor}}^B + D_{\text{fuselage}}^B,
\end{equation}
where $D_{\text{rotor}}^B = [C_{D,rotor},\,C_{D,rotor},\,0.0]^T$, $D_{\text{fuselage}}^B = -C_D\, \text{sign}(v_B)\,v_B^2/m$ with $v_B = R_B^W [\dot{x}\; \dot{y}\; \dot{z}]^T$. The superscripts $B$ and $W$ represent the body and world frames. We note that a more representative model of the aerodynamic effects can be obtained if prior knowledge of these dynamics is available. For instance, the authors in \cite{bauersfeld2021neurobem} combined blade element momentum theory with a neural network to learn the aerodynamic effects. Here, it is assumed that these uncertain dynamics are absent in the nominal model and are to be learnt by the GP and KNODE models. We would also highlight that these simulated uncertain dynamics only manifest in the translational dynamics of the quadrotor and incorporating uncertain rotational dynamics is part of our future work.}

The KNODE model uses a two layer neural network architecture with 64 and 16 hidden units each. A hyperbolic tangent function is used as the activation function. Each of the two trajectories in the training data has a duration of 8s and 4000 data points. The GP model is trained using the scikit-learn package \cite{scikit-learn} and uses a kernel consisting of the product of constant and radial basis function (RBF) kernels. Eighty data points, sampled at regular intervals, are used for training. The inputs for the training data are data samples containing the state-input vector $[\mathbf{x}\;\mathbf{u}]^T$, while the training labels are the differences between the true and nominal state derivatives, \textit{i.e.}, $\Delta\dot{x} := f_{\textbf{gp}}(\mathbf{x},\mathbf{u})-f(\mathbf{x},\mathbf{u})$. With the trained models, simulations are conducted along both types of trajectories to evaluate the prediction accuracy of the KNODE model against the nominal and GP models. To assess generalization performance, the models are tested on trajectories beyond the training data. These include circular and lemniscate trajectories of different radii \revision{and speeds}. Lastly, to test closed-loop performance, the KNODE-MPC framework is simulated on circular and lemniscate trajectories of various radii \previous{\fakest{and compared against a nominal model predictive controller}}\revision{and speeds and is compared against both the nominal MPC and GP-MPC frameworks}. 

\begin{figure}
    \centering
    {\vspace*{0.2cm}\includegraphics[scale=0.35]{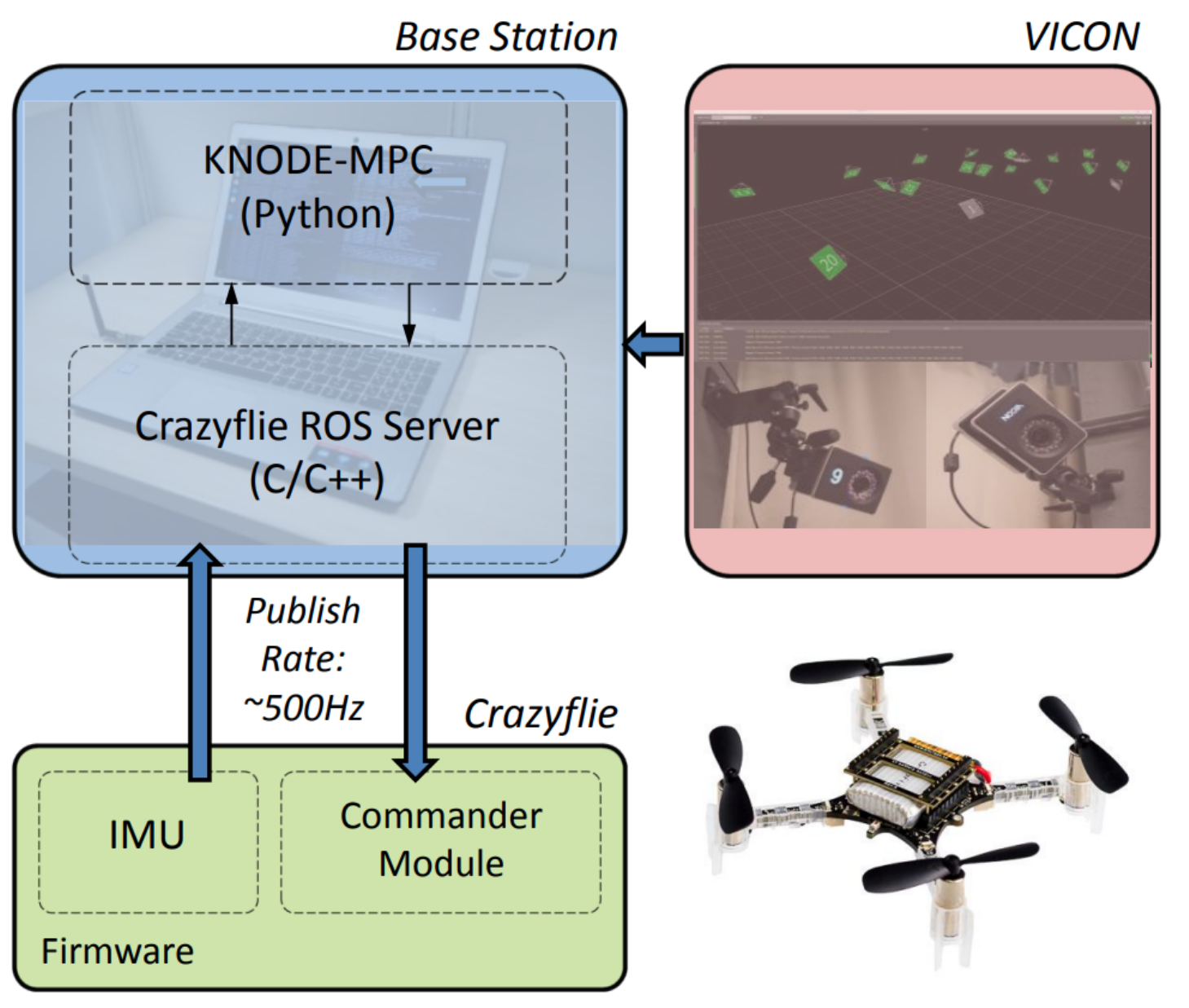}}
    \caption{Schematic of the experimental setup: The Crazyflie experimental platform operating with a Vicon motion capture system. KNODE-MPC publishes control commands through the Crazyflie ROS server and into the commander module on-board Crazyflie.}
    \label{fig:crazyflie}
\end{figure}

\textit{Physical Experiments}: We use the open-source Crazyflie 2.1 quadrotor \cite{Bitcraze} as the experimental platform. An image of a Crazyflie is shown in Fig. \ref{fig:crazyflie}. The Crazyflie, together with motion trackers, weighs 32g and has a size of 9 cm$^2$. The software architecture is developed using the CrazyROS wrapper \cite{crazyflieROS}. A laptop running on Intel i7 CPU acts as the base station and communication with the Crazyflie is done via the Crazyradio PA at \previous{\fakest{a rate of 250Hz}}\revision{an average rate of 500 Hz.} To obtain pose information of the quadrotor, we use a Vicon motion capture system which communicates with the base station via a Vicon bridge and publishes at an approximate rate of \previous{\fakest{88}}\revision{98}Hz. Linear velocities are estimated from the positions obtained from the Vicon, while \revision{accelerations and} angular velocities are measured from the \revision{accelerometers and} gyroscope sensors on-board the quadrotor. This experimental setup is depicted in Fig. \ref{fig:crazyflie}.

To train and validate the KNODE \previous{\fakest{model}}\revision{and GP models}, data is collected by flying the Crazyflie with the nominal MPC framework along circular trajectories of radii 0.5 and 1m \revision{at a speed of 0.5m/s}, each for a duration of \previous{\fakest{76 seconds and 40000 data points. We preprocess the}}\revision{approximately 76 seconds and 39990 data points.} We pre-process the \revision{linear} velocity data using a fifth order Butterworth low pass filter before training. Similar to the simulation setup, the nominal, KNODE-MPC and GP-MPC frameworks are implemented in CasADi and run on the base station. The KNODE model in physical experiments uses a one-layer neural network with 32 hidden units and the hyperbolic tangent activation function. \revision{The GP model in physical experiments uses the same kernel as the one in simulations and it uses eighty equally sampled data points for training.} The frameworks generate three-dimensional acceleration commands and run on top of a geometric controller \cite{mellinger2011}, as well as attitude and thrust controllers within the Crazyflie firmware. For each of the \previous{\fakest{two}}\revision{three} frameworks, we conduct \previous{\fakest{five test runs along circular trajectories of radii 0.5m and 1.25m}}\revision{test runs along circular trajectories of three different radii at a speed of 0.5m/s} to evaluate closed-loop trajectory tracking performance. \revision{We further evaluate velocity generalization performance of the frameworks by conducting tests at a speed different from that in the training data.} We refer the reader to the video in the supplemental materials for a better understanding of the physical experiments.

\subsection{Simulation Results} \label{section:simulations}
\begin{figure}
    \centering
    {\vspace*{0.3cm}\includegraphics[scale=0.305]{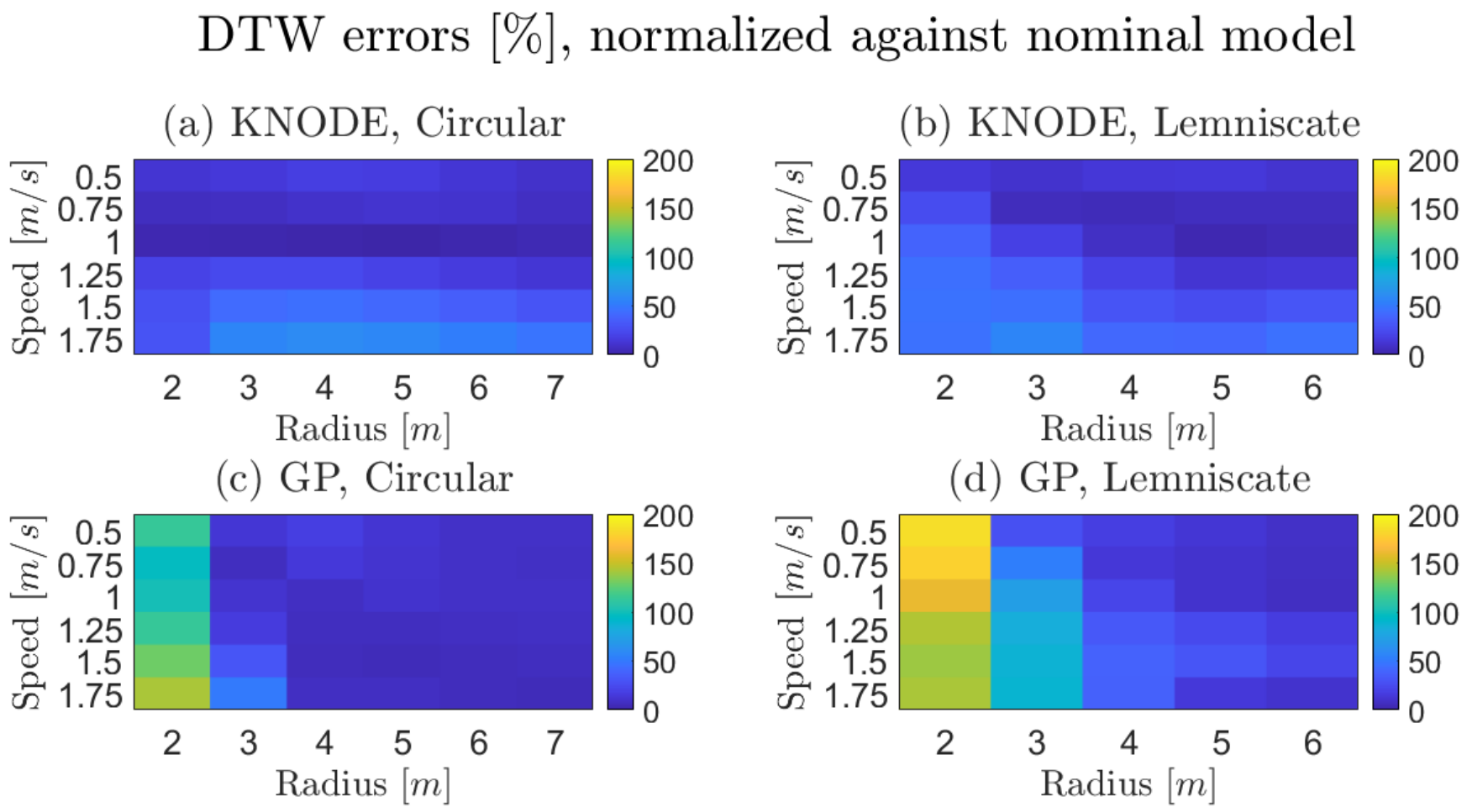}}
    \caption{\revision{\previous{\secondfakest{Statistics}}\secondrevision{Heat maps} of the prediction errors for the nominal, GP and KNODE models, tested with circular and lemniscate trajectories of different radii and commanded speeds. \previous{\secondfakest{The median errors are represented by the line plots, while the ends of the error bars denote the 25$^{\text{th}}$ and 75$^{\text{th}}$ percentiles across trajectories of different radii.}}}}
    \label{fig:open_loop_sim}
\end{figure}

\revision{To establish the accuracy of state predictions, simulations are conducted along circular and lemniscate trajectories of different radii, ranging from 2 to 7 meters and across different commanded speeds from 0.5 to 1.75 m/s. The \previous{\secondfakest{statistics of the}}3D DTW position errors between the predicted trajectories and the true trajectory are plotted in Fig. \ref{fig:open_loop_sim}. \previous{\secondfakest{The line plots represent the median prediction errors, while the ends of the error bars depict the 25$^{\text{th}}$ and 75$^{\text{th}}$ percentiles across trajectories of different radii.}} For clarity, the errors obtained from the KNODE and GP models are normalized by those from the nominal model. \tom{It is observed from Fig. \ref{fig:open_loop_sim} that} both the KNODE and GP models are able to account for the nonlinear aerodynamic drag effects included in the true dynamics and provide accurate predictions. \tom{Considering an overall median error computed over all trajectories across different radii and speeds, both models outperform the nominal model by over 80\%.} Examining both circular and lemniscate trajectories, the accuracy of the KNODE model is \previous{\secondfakest{observed to be}}higher than that of the GP model, with a 19.1\% improvement in the overall median prediction error. Although the GP model provides good predictions in some of the test cases, it has a \secondrevision{relatively} large \previous{\secondfakest{variance}}\secondrevision{variation across trajectories of different radii}, as shown in Fig. \ref{fig:open_loop_sim}\secondrevision{(c) and (d)}\previous{\secondfakest{, implying inconsistent predictions}}. On the other hand, the \previous{\secondfakest{variance}}\secondrevision{variation} of the accuracy of the KNODE model is small across \previous{\secondfakest{all}}trajectories \secondrevision{of different radii and moderate across trajectories of different speeds}, which suggests that the KNODE model is capable of both interpolating within and extrapolating outside of its training domain consistently. For lemniscate trajectories, \previous{\secondfakest{it is observed from Fig. }}\secondfakest{\ref{fig:open_loop_sim}}\previous{\secondfakest{ that}}the KNODE model outperforms the GP model by 45.5\% in terms of the median prediction errors. As lemniscate trajectories are geometrically different from the circular trajectories \tom{in the training data}, these results further ascertain the generalization capability of the KNODE model.}

\begin{figure}
    \centering
    {\includegraphics[scale=0.295]{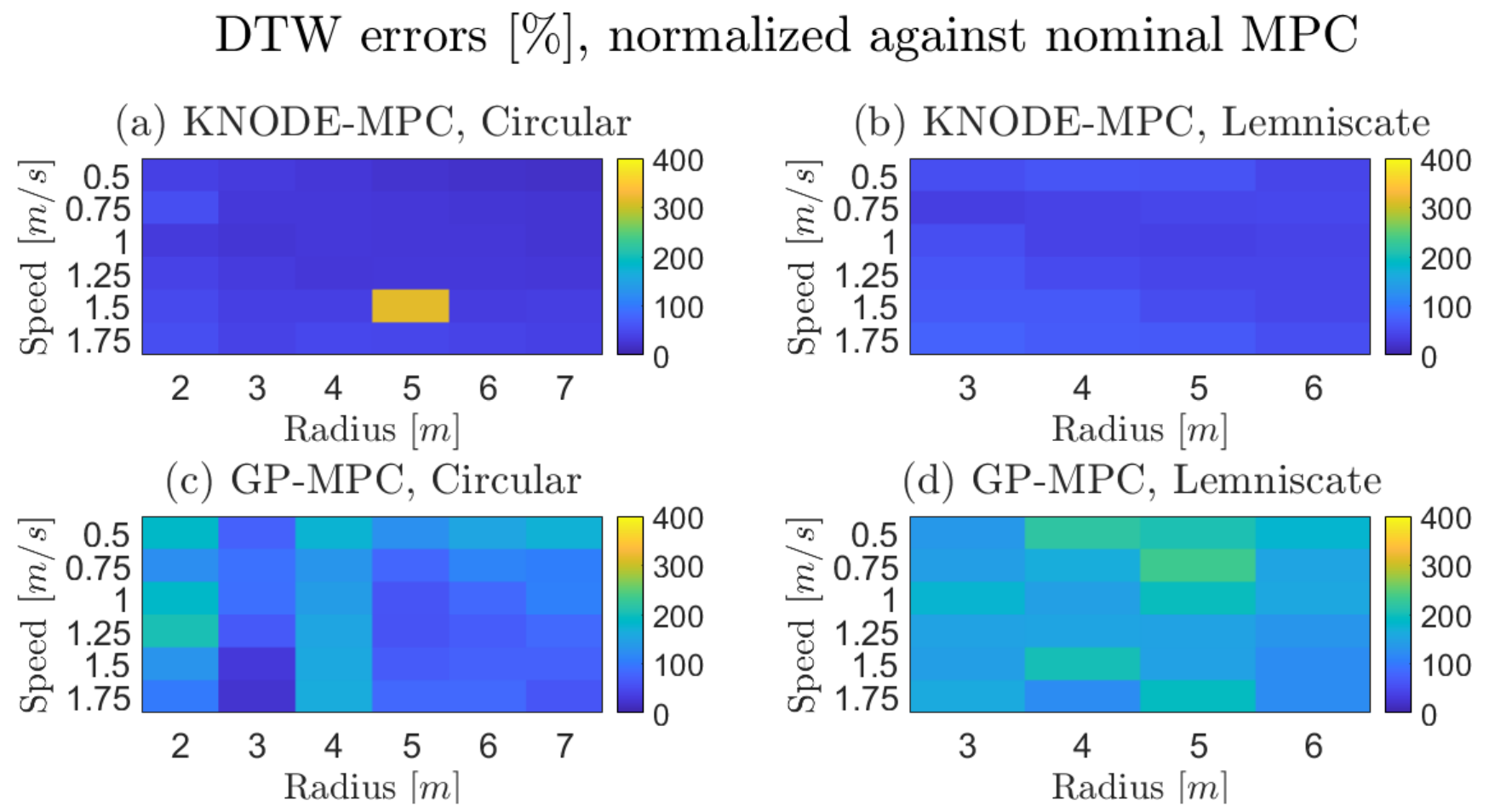}}
    \caption{\revision{\previous{\secondfakest{Statistics}}\secondrevision{Heat maps} of the closed-loop trajectory tracking errors under the GP-MPC and KNODE-MPC frameworks, along circular and lemniscate trajectories of different radii and speeds. \previous{\secondfakest{The median errors are represented by the line plots, while the ends of the error bars denote the 25$^{\text{th}}$ and 75$^{\text{th}}$ percentiles across trajectories of different radii.}}}}
    \label{fig:closed_loop_sim2}
\end{figure}

\revision{Next, closed-loop simulations using nominal MPC, KNODE-MPC and GP-MPC are conducted along circular and lemniscate trajectories of different radii and commanded speeds. Results are plotted in Fig. \ref{fig:closed_loop_sim2}. \secondrevision{The trajectory tracking errors for KNODE-MPC and GP-MPC are normalized by those obtained with nominal MPC.} Compared against nominal MPC and GP-MPC, KNODE-MPC improves the median trajectory tracking errors by 60.2\% and 67.8\%, computed across all test cases. \secondrevision{Although an anomaly is observed for the trajectory under KNODE-MPC at radius 5m and speed of 1.5m/s, t}\previous{\secondfakest{T}}he smaller \previous{\secondfakest{variances}}\secondrevision{variation} across the\previous{\secondfakest{se}} \secondrevision{remaining} test cases demonstrate\secondrevision{s} the consistency and effectiveness of the integrated KNODE-MPC framework in the presence of uncertain dynamics. This further demonstrates the generalization capability of the KNODE-MPC framework, since the models are only trained with data at radii of 3 and 6m, with a commanded speed of 1m/s. From these results in Fig. \ref{fig:closed_loop_sim2}, we note that the integration of the GP model described in Section \ref{section:evaluation} with MPC does not necessarily yield better closed-loop performance, even though the open-loop state predictions are sufficiently accurate, as shown in Fig. \ref{fig:open_loop_sim}. On the other hand, the KNODE model, which is more consistent in state predictions, provides better closed-loop performance, upon integration with MPC.}

\begin{figure}
    \centering
    {\includegraphics[scale=0.295]{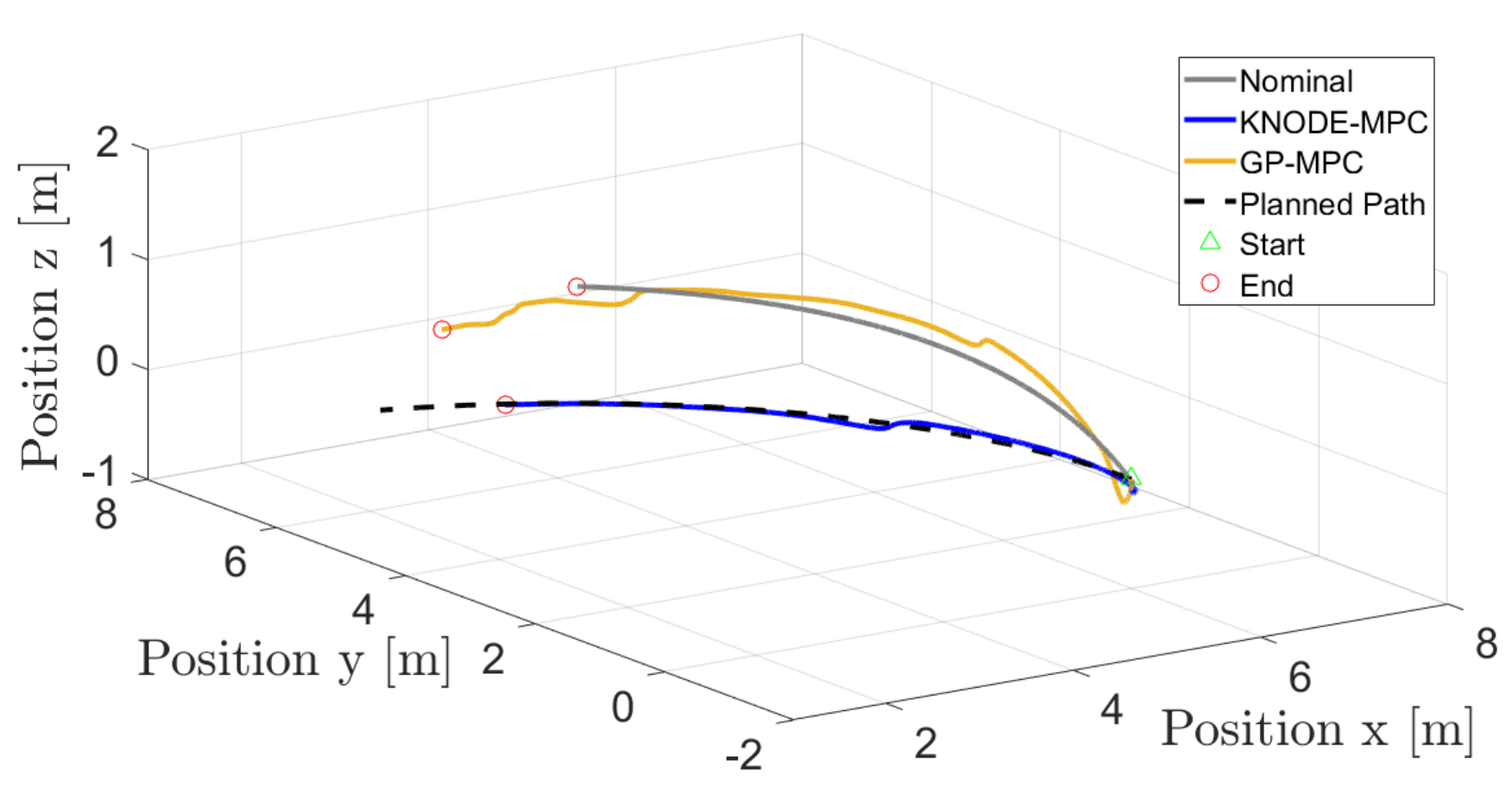}}
    \caption{\revision{3D trajectories under the nominal MPC, GP-MPC and KNODE-MPC frameworks, on a segment of a circular trajectory of radius = 6m and at a speed of 1m/s.}}
    \label{fig:closed_loop_traj}
\end{figure}

Fig. \ref{fig:closed_loop_traj} provides a \revision{qualitative} comparison between the closed-loop trajectories under the \previous{\fakest{two}}\revision{three} frameworks. Since the model in the nominal MPC framework does not account for residual or uncertain dynamics, the quadrotor trajectory deviates from the planned trajectory, with position errors accumulating over time. On the other hand, the KNODE model compensates for the residual dynamics and allows the quadrotor to follow the planned trajectory more closely, \revision{while outperforming the GP model.}

\tom{Additionally, we performed experiments to quantify the sample efficiency of training the KNODE model in simulation. Five KNODE models are trained on different fractions of the training data specified in Section \ref{section:setup}, and the resulting models are evaluated for closed-loop DTW trajectory tracking errors on circular trajectories with radii of both 3m and 4m. The errors are normalized against those from nominal MPC. Results are shown in Table \ref{tab:sample_complexity}. The errors decay as the fraction of data used for training approaches 1 for both radii. Furthermore, it is observed that the training time scales linearly with the number of training data.}

\begin{table}[!t]
\vspace*{0.4cm}\caption{\revision{Sample complexity of KNODE-MPC.} \label{tab:sample_complexity}}
\centering
\begin{tabular}{|c||c|c|c|c|c|}
\hline
Training Data Fraction & 1/16 & 1/8 & 1/4 & 1/2 & 1\\
\hline
DTW error $[\%]$, radius=3m & 102.7 & 76.6 & 63.4 & 42.0 & 26.2 \\
\hline
DTW error $[\%]$, radius=4m & 101.0 & 85.0 & 53.5 & 29.5 & 28.9\\
\hline
Training Time [s] & 876 & 955 & 1103 & 1426 & 1890 \\
\hline
\end{tabular}
\end{table}

\begin{figure}
    \centering
    {\vspace*{0.2cm}\includegraphics[scale=0.28]{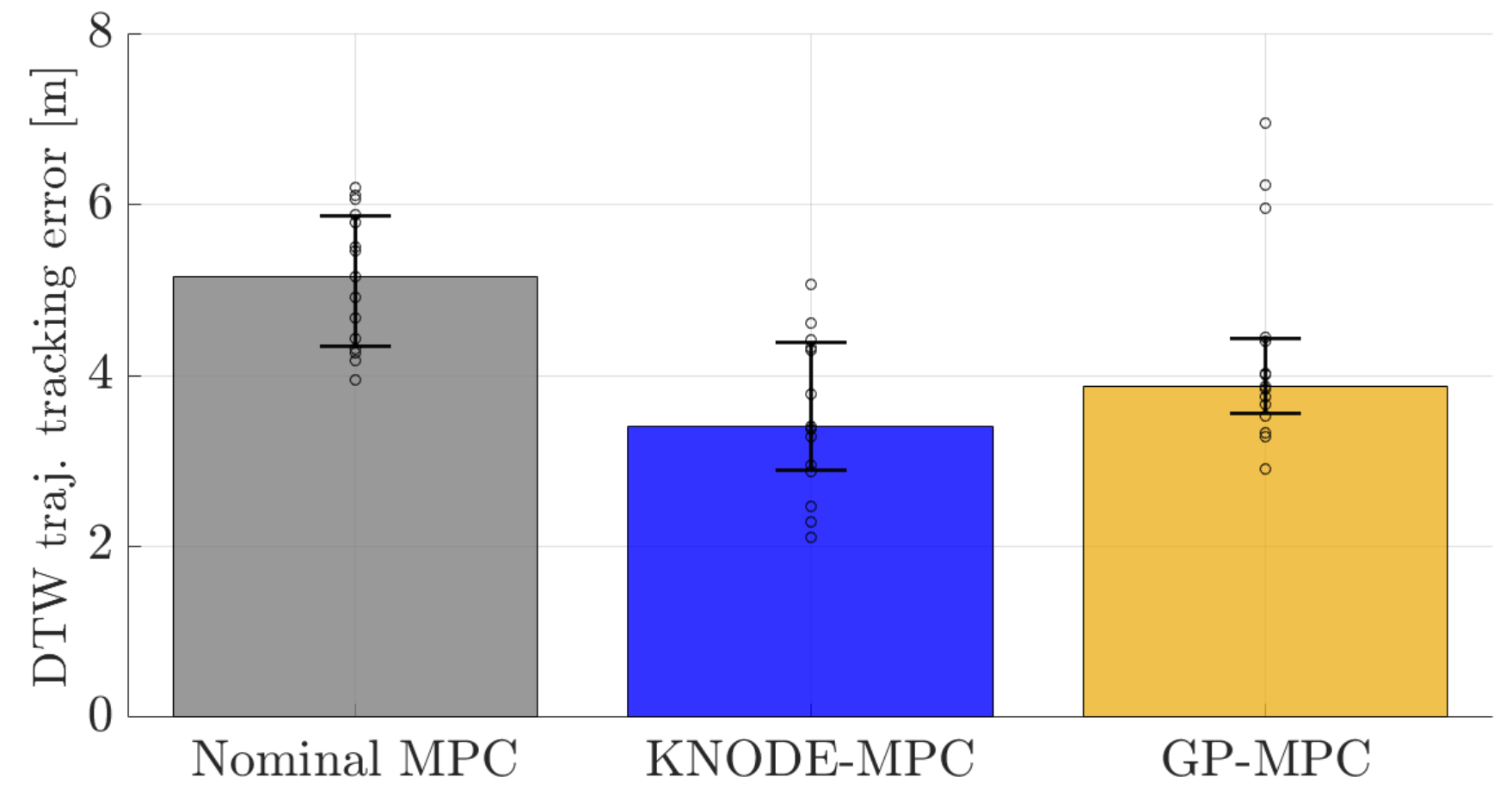}}
    \caption{\revision{Performance on training trajectory: Statistics for nominal MPC, KNODE-MPC and GP-MPC along a circular trajectory of radius 0.5m across 15 runs for each framework. The top of the bars represent the median and the ends of the error bars depict the 25$^{\text{th}}$ and 75$^{\text{th}}$ percentiles. The trajectory tracking errors for each of the runs are illustrated with circles.}}
    \label{fig:expt_training_dat}
\end{figure}

\textcolor{black}{\subsection{Physical Experimental Results}} \label{section:experiments}
\revision{To verify closed-loop performance, all three frameworks, namely nominal MPC, KNODE-MPC and GP-MPC, are tested in physical experiments. First, the frameworks are tested with a circular trajectory of radius 0.5m at a speed of 0.5m/s. This trajectory is part of the training data and improvements over nominal MPC are to be expected for both KNODE-MPC and GP-MPC. We conduct 15 runs for each framework, with a total of 45 runs. Statistics of the planar DTW trajectory tracking errors are plotted in Fig \ref{fig:expt_training_dat}. The colored bars represent the median across 15 runs for each framework and the ends of the black error bars denote the 25$^{\text{th}}$ and 75$^{\text{th}}$ percentiles. The results obtained from all runs are plotted with circles. Comparing the median errors, we observe a 34.1\% improvement for KNODE-MPC over nominal MPC and 12.4\% improvement over GP-MPC.}

\begin{figure}
    \centering
    {\vspace*{0.2cm}\includegraphics[scale=0.295]{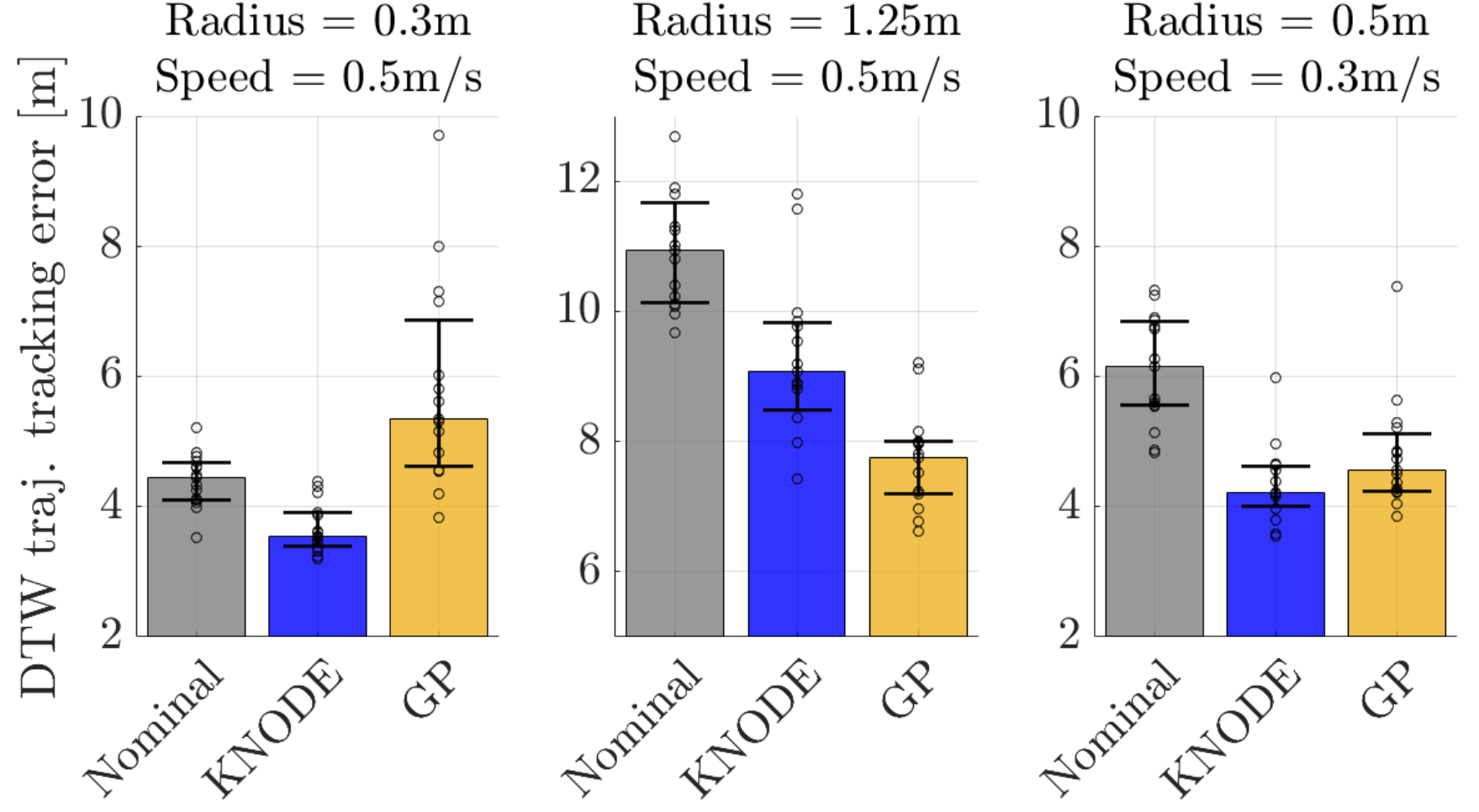}}
    \caption{\revision{Generalization performance: Statistics for the three frameworks under three scenarios that are used to ascertain generalization ability. For each of these scenarios, 15 runs were conducted for each framework. The top of the bars represent the median and the ends of the error bars depict the 25$^{\text{th}}$ and 75$^{\text{th}}$ percentiles. The DTW errors for each of the runs are illustrated with circles.}}
    \label{fig:expt_gen}
\end{figure}

\revision{Next, the generalization ability of KNODE-MPC is investigated using trajectories beyond its training dataset. Three test scenarios are considered. For the first two scenarios, the planned trajectories have radii of 0.3m and 1.25m, which differ from training trajectories, which have radii of 0.5m and 1m. These trajectories have a commanded speed of 0.5m/s. In the third scenario, a speed of 0.3m/s is commanded and this is to investigate generalization in velocity for KNODE-MPC and GP-MPC. A total of 135 test runs are conducted, fifteen runs for each framework and for each of the three scenarios. Fig. \ref{fig:expt_gen} depicts the statistics of the planar trajectory tracking errors for these test runs. KNODE-MPC outperforms nominal MPC in all three cases with an overall improvement of 21.0\%, in terms of the combined median trajectory tracking error. This combined median error is computed by considering errors from all 135 runs, which are illustrated with circles in Fig. \ref{fig:expt_gen}. Notably, KNODE-MPC also performs better than GP-MPC across these scenarios by 14.6\%, in terms of the overall median trajectory tracking error. Analyzing the results for each of these scenarios, it is observed that the performance of KNODE-MPC is more consistent than that of GP-MPC. In particular, even though GP-MPC performs better than nominal MPC for the second and third scenarios, it performs worse in the first scenario. On the other hand, KNODE-MPC performs better than nominal MPC under all three scenarios.} The differences in performance improvement between simulation and physical experimental results can be attributed to the different control architectures between the two setups. In particular, the KNODE-MPC framework in simulations generate control commands to actuate the quadrotor directly, while in physical experiments, it runs on top of the geometric controller and low-level controllers within the firmware, and it does not have direct control over the quadrotor dynamics.

\section{Conclusion and Future Work}

In this work, we presented the KNODE-MPC framework with an application to quadrotor control. The knowledge-based, data-driven model, KNODE, provides a more accurate representation of the system dynamics, as compared to a state-of-the-art GP model. By incorporating KNODE into the MPC framework, the predictive controller generates optimal control commands that yield better closed-loop performance. This framework is tested extensively in simulations and physical experiments. Results show significant improvements in terms of state predictions, as well as closed-loop trajectory tracking. Given its flexibility, the framework can be applied to robotic systems operating in complex and uncertain environments such as marine robots or robotic teams. In the near future, we would like to extend this framework to an online learning setting, where the KNODE model can be updated and refined in real-time, which can further improve the performance of closed-loop control tasks. 

\bibliographystyle{IEEEtran} 
\bibliography{IEEEabrv,refs} 

\addtolength{\textheight}{-12cm}   





\end{document}